# The Duality of Data and Knowledge Across the Three Waves of AI

Amit Sheth
AI Institute
University of South Carolina

Krishnaprasad Thirunarayanan
Computer Science & Engineering Department
Wright State University

We discuss how over the last 30 to 50 years, Artificial Intelligence (AI) systems that focused only on data have been handicapped, and how knowledge has been critical in developing smarter, intelligent, and more effective systems. In fact, the vast progress in AI can be viewed in terms of the three waves of AI as identified by DARPA. During the first wave, handcrafted knowledge has been at the center-piece, while during the second wave, the data-driven approaches supplanted knowledge. Now we see a strong role and resurgence of knowledge fueling major breakthroughs in the third wave of AI underpinning future intelligent systems as they attempt human-like decision making, and seek to become trusted assistants and companions for humans. We find a wider availability of knowledge created from diverse sources, using manual to automated means both by repurposing as well as by extraction. Using knowledge with statistical learning is becoming increasingly indispensable to help make AI systems more transparent and auditable. We will draw a parallel with the role of knowledge and experience in human intelligence based on cognitive science, and discuss emerging neuro-symbolic or hybrid AI systems in which knowledge is the critical enabler for combining capabilities of the data-intensive statistical AI systems with those of symbolic AI systems, resulting in more capable AI systems that support more human-like intelligence.

**ROLE OF DATA AND KNOWLEDGE IN AI**

The roles of data and knowledge in AI have been extensively debated. Knowledge has been synthesized from data in many different ways, or manually codified to model the language we use or the way the world around us works, to enable perception, querying, prediction, and explanation. We believe that the approach to acquiring knowledge and the form of knowledge should fit the context of use and application and there is no one-size-fits-all approach.

In the first wave of AI in the 1980s and early 1990s, the ability to perform symbolic computation over and beyond numeric processing was considered a sign of intelligence. Subsequently, symbolic encoding of domain knowledge in logic and using that to reason with data became the primary approach. In fact, for knowledge representation and reasoning, two separate camps and approaches evolved: a declarative approach where knowledge captured the way the world worked and was separated from how it gets used and a procedural approach where the knowledge was intertwined with the way it was used. In other words, in the former case, explicit knowledge acquisition can be cleanly separated from the multitudes of ways it was used, while, in the latter case, the knowledge was implicit and an integral part of the application-specific code. While the former is beneficial from a broad reuse perspective, the latter became necessary for building efficient implementations in practice using heuristics developed and incorporated with the domain knowledge into the application code [Neats & Scruffies].

In this century, during the second wave of AI, big data, including large volume, a variety of data forms, and velocity representing changing information about an entity or event, became available. As a result, a manual approach to symbolic knowledge acquisition did not seem viable or scalable. So different forms of knowledge and corresponding approaches to acquire that knowledge automatically from data became popular. The pendulum swung from the primacy of manual curation of knowledge to the primacy of using data to glean knowledge, and the progression from knowledge-based approaches based on logic, to statistical approaches based on probability theory, to model fitting based on neural networks and deep learning. The approaches based on formal semantics aligned with explicit knowledge while those based on informal semantics aligned with implicit knowledge [Sheth et al 2005]. And the AI practitioners got fragmented into several different camps with extreme views on the efficacy of one approach to the exclusion of the others, which, in our opinion is neither tenable nor productive. We advocate a middle ground that is both inclusive and unifying where we use the best approach from each camp for the application it fits and hybridizes multiple approaches when a complex multi-faceted application demands it [Dominguez 2017].

In this paper, we review, at a high-level, existing approaches to knowledge acquisition, representation, and reasoning, and discuss their pros and cons and the need to develop approaches to combine them suitably to get the best from each, and to the benefit of the others. We contend that hybridizing knowledge-based approaches with deep learning will eventually promote generality and robustness. While we are not advocating anything revolutionary, the recent polarization of various camps seems to suggest that revisiting George Santayana's warning about "*Those who do not remember the past are condemned to repeat it*" is wise.

**LOGIC-DRIVEN SYSTEMS**

A variety of logic-based formalisms have provided and continue to provide a foundation for building knowledge representation and reasoning systems that underlie AI.

**Deductive Reasoning**
Early AI involved abstracting the world's knowledge in terms of objects, functions, and relationships, and using logic to perform deductive reasoning to make explicit information that is implicit. The needed expressive power and efficiency of computation dictated the choice between the continuum of propositional logic, relational algebra, datalog, Horn logic, and first-order logic, and their expressive extensions [Logic Theory 2018]. While these formalisms were originally introduced to formalize classical mathematics and discover "truths in nature,", these formalisms have enabled encoding of different kinds of knowledge about the world around us including taxonomic (inheritance), part-of, causal, and situational, that related to planning and action, and even meta-knowledge about cognitive processes. These representational formalisms provide notational efficacy and deductive capabilities that can be used to make explicit what is implied, by querying the current state and predicting the future states. Human-curation is possible for high-quality linguistic and situational knowledge but manual acquisition is labor-intensive. These monotonic formalisms are appropriate for formalizing "definite" or "certain" knowledge. Monotonicity refers to the property that if we expand the set of facts, the set of conclusions drawn from the facts will also expand. But, in contrast with classical mathematics that can be formalized using monotonic first-order logic, common-sense reasoning requires us to draw conclusions and act in the face of incomplete knowledge.

**Non-Monotonic Reasoning**
One of the key aspects of common-sense reasoning is our ability to make assumptions to enable action, derive/observe consequences, and then backtrack if the actions based on these assumptions are discovered to be untenable. Non-monotonic reasoning formalisms were introduced to represent exceptions and reason in the presence of incomplete knowledge. These have their humble beginnings in semantic networks and truth-maintenance systems used for representing real-world knowledge and common-sense reasoning. Non-monotonicity refers to the property that if we expand the set of facts, the set of conclusions drawn from the updated facts may not contain the earlier conclusions. The incompleteness arises because it is not practical to represent all the information, or that the information is not readily available, or is discovered only as a consequence of a failure of action in the real world, or when a contradiction is encountered. For instance, we sometimes derive negative information in the absence of explicit support for positive information (closed-world assumption). Non-monotonicity stems from the need to update assumptions that require the previous conclusions to be overridden, retracted, and revised.

**Integrating and Interleaving Deductive and Abductive Reasoning**
Deductive reasoning enables reasoning from causes to effects, from events to consequences, and from observations to predictions utilizing appropriate domain knowledge about nature (such as about the properties of objects around us, about motion and forces, about health/medicine using science) or engineered artifacts (such as using information about digital technology and engineering design). Abductive reasoning enables reasoning from effects to causes to treatments and from observations to hypotheses and explanations. Each of these steps can be viewed at different levels of abstraction. For example, sensing and performing actions using sensors and actuators may demand interleaved execution of deduction-abduction perception cycle mediated by structured domain knowledge, where the sensed state of the world may require further selective observations for disambiguation [Henson et al 2011, Thirunarayan and Sheth 2015]. For another example, consider diagnosing and treating a disease that may require going from symptoms to potential diseases to confirmatory tests to medications and remedial measures. Further, logical formalisms can be used to represent the spatio-temporal context and formalize spatio-temporal reasoning and trend analysis [Walter and Zakharyaschev 2003]. The key aspect of these representational formalisms is our ability to formally capture the structure and the behavior of the objects and systems around us and audit the reasoning.

**Semantic Web Formalism**
The Web 3.0 ecosystem enabled one to bridge coarser information retrieval with more fine-grained data retrieval to hybridize human and machine accessibility. Semantic Web (Web 3.0) formalisms and associated technologies evolved to ensure scalability, universal access, and efficient machine processing of logic-based descriptions [Berners-Lee et al 2001, Sheth and Thirunarayan 2012]. Universal Resource Locators (URLs) provided a global namespace for referring to entities unambiguously, Resource Description Formalism (RDF) provided a formal means to associate metadata with an entity to describe its characteristics, Web Ontology Language dialects (OWL) provided a means to represent and reason with knowledge trading expressiveness with computational tractability, and SPARQL provided SQL-like easy-to-use but richer query language.

**TRADITIONAL DATA-DRIVEN SYSTEMS**

Early data-driven systems attempted to capture correlations, statistical regularities, and patterns in data and codify them as rules for wide-spread use.

**Data Mining**

Data mining has been used to glean patterns and trends hidden in situational or transactional data to synthesize associational knowledge that can shed light on "local" correlations, behaviors, and anomalies. The explicit associational knowledge can summarize raw data to uncover hidden insights, and ultimately enable decisions and actions. In contrast with logic-based approaches where we formalize known relationships to reason with data, data mining approaches attempt to assimilate observational data to uncover and abstract relationships implicit in the data. The former is used to capture certain/definite knowledge, while the latter discovers statistical knowledge.

**Probabilistic Systems: Complementing Logic-Driven Systems**

The knowledge that can be acquired about some domains/applications, or gleaned from data can be incomplete for a number of reasons, and we need suitable representational formalisms to capture the intrinsic nature of the domain and the situation to reason with them. For instance, the incompleteness may be (i) the result of not having all the relevant information, yielding only tentative conclusions that can be overridden in due course (when exceptions are recognized or temporal evolution is accounted for), requiring non-monotonic extensions to logic, or (ii) due to the stochastic nature (at the mercy of random events) but exhibiting discernible trends over a large number of observations capturable through probabilities and probabilistic reasoning, or (iii) due to the fuzzy nature of the linguistic concepts (convenient and natural for human communication) to be formalized using fuzzy logics. Each different source of incompleteness should be handled according to its intrinsic nature, which is why numerous representational languages and reasoning mechanisms have been proposed and developed. As is evident from the above examples, while some associations are statistical in nature that can be gleaned from the data, others have conceptual, taxonomic, cognitive, or linguistic origins requiring manual curation.

In its simplest incarnation, Bayesian reasoning, founded on the rules of probabilities and conditioning, can be used to summarize observational data to capture "events-consequences" or "causes-effects" relationships and can be used to infer most likely causes from a collection of observed symptoms. Bayes rule formalizes reasoning with sets and proportions -- overlapping and disjoint sets that can be visualized using Venn Diagrams. This formal basis promotes clarity of understanding, providing a foundation for rigorous reasoning reduced to calculation, and has the potential to elucidate unintuitive consequences of some diagnostic problems (such as providing an argument for why a positive result from a diagnostic test for a rare disease still deserves a second opinion). Bayesian reasoning can also provide a means to refine probability estimates of an event from observational data. That is, we can reason from event probabilities to its consequences as well as from consequences to event probabilities. The latter statistical reasoning has applications ranging from medical diagnosis and fault diagnosis to inferring trustworthiness of agents and systems such as sensor networks and recommender systems.

Probabilistic graphical models (PGMs) (including Bayesian Networks) marry probability theory and graphical structures to develop representational language with a clean semantic basis for interpretation and enabling graph-based reasoning algorithms for efficiency. PGMs can be used to represent both qualitative dependencies and quantitative parameters gleaned from data. PGMs can incorporate known qualitative relationships and quantitative strength of associations. In other words, PGMs provide a means to incorporate human-curated declarative knowledge as well as qualitative associations gleaned from the data while customizing quantitative parameters for each situation based on the corresponding observational data (see for example [Anantharam et al 2016]).

While the accuracy of the estimated probabilities depends heavily on the representativeness of the sampled data, probabilistic reasoning can yield robust conclusions if we can use probability values to rank different alternatives reliably as long as the errors in the estimates do not interfere with the ordering. This robustness aspect of probabilistic reasoning has been critical to the practice of (Naive Bayes) classification, (model-based) recognition, and planning tasks in the real world.

**NEURAL NETWORKS AND DEEP LEARNING**

Neural networks have been used to learn complex associations from large amounts of data, to perform various downstream tasks ranging from classification and recognition to language translation, summarization, and generation. In the beginning, the two-layer perceptron model served as the basis for learning a linear classifier. Subsequently, multi-layer perceptron models were trained to learn non-linear functions of the inputs from large amounts of data. Conceptually, training a neural network involves solving an optimization problem to fit a "least-error" curve/surface, or equivalently, learn a best-fit multivariable function given a finite set of input-output pairs. While the neural network so trained can interpolate and extrapolate from the training set, it does not provide an explicit description of the association that is accessible to a human in a form that can be scrutinized and analyzed. **This black-box nature of neural networks can hamper its broader adoption in some critical application domains where a human-in-the-loop is necessary to rationalize actionable decisions to inspire confidence.**

To better situate neural networks among the information representational frameworks and for carrying out machine learning tasks, lets us explore its pros and cons. Neural networks can be trained efficiently on large datasets using deep learning algorithms -- supervised approaches using labeled examples (such as commonly used in tasks involving image, video, or audio data) or unsupervised approaches where the context in which the data element appears defines the semantics of the data element (such as in word-embedding based vector representation of text fragments). The key benefit of neural network architectures (ranging from Feedforward Networks, Convolution Neural Networks and Recurrent Neural Networks to transformers, advanced sequence models, and General Adversarial Networks) and associated deep learning algorithms is their ability to automatically extract significant/relevant features required for the task from the raw data. However, this assumes that the dataset used for training is representative as well as unbiased for the purposes of the task [Suresh and Guttag 2019], which is often not the case in practice.

The data available in the wild may not be representative or may contain anomalies or gaps with respect to the features that are essential for decision making. An unbiased subset of the data is necessary to avoid the "sins of the past" baked into the available (historical) data. For instance, for social good applications, the data used and the decisions made on its basis may be distorted because of unequal access to data sharing capabilities of the affected populace, or skews in the available data due to socio-economic, demographic (e.g., age, race, and gender), geographic, and historical factors, or due to hidden limitations in collection and aggregation strategies and platforms. The quality and the reliability of the conclusions we can derive, through the application of data-driven techniques on the digital traces we leave behind us, are as good as the fidelity and generality of the data we feed in. **Key limitations of using neural networks and deep learning include the difficulty in characterizing hidden biases and quality issues in data, making it vulnerable to spurious/erroneous correlations/associations. This lack of representative and unbiased data can hamper the adoption of deep learning algorithms in critical applications that require guarantees, transparency, and accountability.**

A benefit of using multi-layered neural networks trained on a large dataset for a generic task (such as image classification using ImageNet) is their ability to be customized for a specific task by freezing all but the last few layers and retraining the last layers for the specific task. This enables the reuse of existing investment in training the neural network for the general problem to obtain broadly useful high-level features, and then using transfer learning to adapt these trained networks to specific tasks by selecting/weighting relevant features in a scalable -- incremental and distributed --- manner. Heterogeneous transfer learning can be used to mediate between domains and modalities that have different feature spaces [Day and Khoshgoftaar 2017].

The deep learning algorithms have been versatile and effective in ingesting data of different modalities (ranging from images and video to audio and text) and performing classification, recognition, translation, and generation tasks, but the "logical mechanism" by which they accomplish these tasks so admirably are opaque. **While deep learning enables one to perform empirically defined tasks well in the aggregate, it is not conducive to accomplish general or human-like intelligence tasks accurately because we are unable to scrutinize and programmatically exploit the learned representations to audit it to provide formal guarantees in the conclusions derived.**

Let us review, at a high-level, the nature of typical applications in which deep learning excels as well as performs poorly, to rationalize the underlying reasons and "mechanisms" for their success as well as to remedy the limitations. The deep learning algorithms excel in automatically but opaquely uncovering patterns and trends in a variety of data that enable data compression by detecting and encoding redundancies during training. While the decoding process can enable good performance on various downstream tasks including reconstructing an approximation to the original data, it is difficult to verbalize the essential features that are abstracted to serve as an explanation for the conclusions (albeit with some limited success in applications such as image recognition or word analogies). **The deep learning algorithms excel in automatically but opaquely uncovering semantic equivalence and subsumption relationships based on the similarity of usage contexts detected and encoded during training. This provides a limited means to impose hierarchical organization (e.g., IS-A, HAS-A). However, this is not robust with respect to reliably uncovering synonymy, polysemy, part-of/part-whole/has-a, and other labeled relationships in general, requiring manually curated resources such as WORDNET for formal arbitration of linguistic knowledge or UMLS for biomedical knowledge.**

**Low-level Perception vs High-level Reasoning**

Neural network architectures and deep learning algorithms are powerful mechanisms for low-level perception as they can extract features from low-level sensor data such as images, videos, and audio streams, and from textual data available from sources such as news documents, scientific literature, and social media streams. These features can be associated with entities, formalized and represented as facts in a representational language, and, in conjunction with codified domain knowledge capturing relationships, can be used for high-level reasoning underlying querying, question-answering, decision-making, and action-planning. **Formally describing the features in terms of their low-level manifestation in sensory or expressed data using knowledge-based formalisms explicitly is, at best, highly cumbersome, and at worst, practically impossible. However, a variety of machine learning techniques (supervised, semi-supervised (active and transfer learning), and unsupervised) are available to extract features automatically. One can build a knowledge-based system that uses these features for transparent decision making and explanation generation, in contrast with the state-of-the-art neural network models that are black-box and not human-auditable.** For example, there are image recognition examples where a human perceives an image one way but the same image is grossly mislabeled by a neural-network classifier. And it is so egregious that one can train ordinary networks systematically using gradient ascent methods or train Generative Adversarial Networks to create hard-to-detect fakes!

**The fundamental problem is that neural networks can learn anomalous discriminating features that are adequate to separate training samples belonging to different classes without necessarily also learning the characteristic features that can reliably distinguish members of the different classes when used in practice.** For instance, how do we know that a classifier that separates wolf images from dog images is not learning how to separate a "snowy" background from a "normal" background? This is reminiscent of the correlation vs causation

debate. **To the extent that knowledge-based systems can declaratively specify and exploit "causative" features characterizing classes, in preference to "correlated" features implicitly learned from the training samples, their integration to get the best of both the worlds will create a powerful and reliable system. An attractive option is to use a two-stage representation and reasoning system that uses neural networks and deep learning algorithms for low-level perceptual tasks while using a knowledge-based system built on top for high-level reasoning and decision making.** See Figure1.

Concretely, the lower-level (first stage) can sense and collect data, and use neural networks and deep learning to extract, abstract, and classify features to recognize primitive entities and relationships. These can be codified by creating and/or instantiating a knowledge graph: nodes representing entities and edges representing relationships. The upper-level (second stage) formalizes symbolic reasoning embodying deductive, abductive, non-monotonic, default, and probabilistic reasoning over the knowledge graph in terms of graph patterns and paths to obtain decisions and actions. Further, iterative and interleaved top-down bottom-up reasoning that underlies human-pattern recognition such as what is required to process and interpret fruit-faces can be accomplished in the second stage [Palmer 1975].

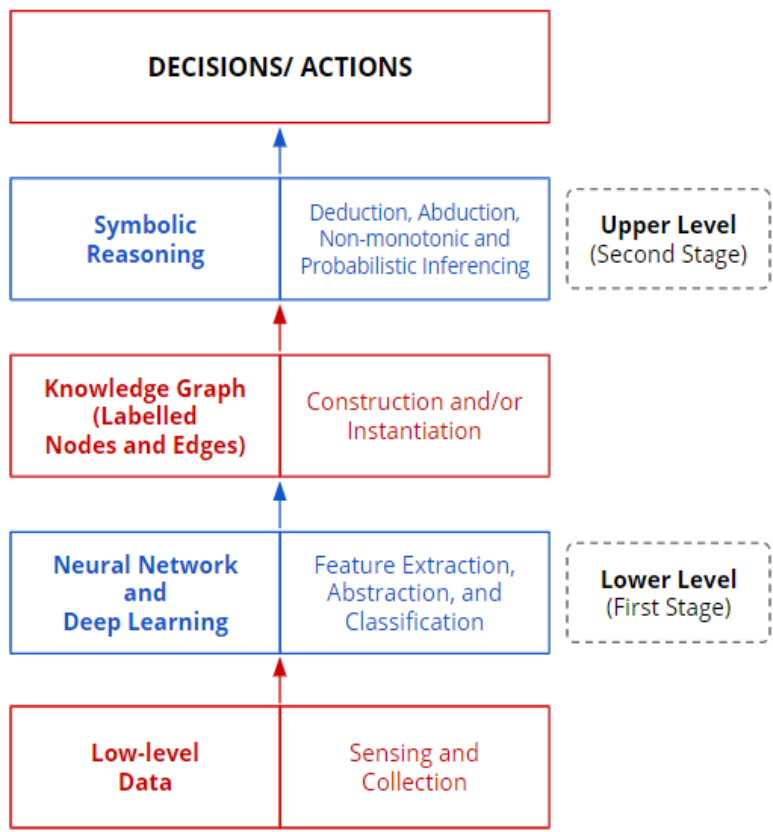

Figure 1: Layered hybrid system using neural networks and deep learning for perception with knowledge-based reasoning system for decision making

**Declarative Knowledge and Exceptions: Representing Regularity Explicitly**

There have been provocative claims that deep learning systems are the be-all and end-all for knowledge acquisition, representation, reasoning, and deployment. **What this argument is missing is the rationale for learning and rediscovering declarative knowledge from the available data that we have already created, verified, and curated over decades (especially in the absence of reliable techniques to formulate those "invariants" symbolically from data)?** (Why not have the cake and eat it too!) Further, it is unclear if the data generated from the validated declarative models (such as those capturing the laws of nature) can be used to automatically reverse engineer the symbolically expressed models (invariants) that created them, without human inspiration, insight, or guidance (such as what was done by scientists of the yore) without imposing "non-trivial" constraints on the data to be used. And even if we can, why reinvent the wheel? We believe that finding means to integrate declarative knowledge with neural networks, each contributing its strengths, is a pragmatic and viable path to explore and develop richer systems. Further, automatic knowledge acquisition may be able to capture regularities and associations for coarse-grain structures (superclasses) but may fail to do the same for fine-grained structures (subclasses) due to data imbalance and sparsity, providing an inadequate basis for refinement. The human-in-the-loop mediation (via active learning) to properly learn hierarchical structures may be the analog of incorporating declarative knowledge. In other words, capturing the duality of data and knowledge, mediated by neural networks and deep learning to comprehend data, seems to hold promise for developing practical intelligent systems that are interpretable, explainable and auditable.

**Overcoming Sparsity and Bias: Representing Anomalies Explicitly**

**When studying issues related to how well available data represents real-world scenarios, we can distinguish between (i) *sparse data* about an event or situation that occurs only rarely, (ii) *absent data* about an event or situation that has not yet occurred but can occur, and (iii) *skewed/biased data* about an event or situation because of incompleteness in its generation or capture. In all these cases, the use of external knowledge seems fundamental to overcome the inherent limitations in the available data and to support reliable reasoning and decisions.** Specifically, we can address sparse data challenges by incorporating expert-curated knowledge about the rare but known cases to improve the effectiveness of the eventual decisions (such as for the diagnosis and management of rare diseases). On the other hand, the only way to deal with absent data is to specify declaratively and/or simulate unrealized conditions, to determine eventual decisions (such as when dealing with modes of failures of an autonomous vehicle or planning for an assortment of mobile objects that may be detected by traffic cameras). In the case of skewed data, characterizing and formalizing sources of bias and incompleteness (such as data that is only from social media platforms or is about certain demographics) can enable the determination of whether a conclusion is warranted or unwarranted.

**Semantics-respecting integration of Deep Learning Embedding Models: Elevating Distributional Semantics**

Embedding models, such as Word2Vec, Glove, BERT, and a host of other domain-specific variants, were developed to glean and represent (through unsupervised learning) the semantics of words using the contexts in which they appear in a large text corpus (such as Wikipedia and PubMed). Word embeddings are usually computed using neural network-based deep learning architectures and are typically a 300-dimensional vector of latent dimensions that capture the senses in which a word can be used. This permits reasoning about word equivalences in a way that is usually complementary to using a manually curated thesaurus for the same purpose. However, two critical issues have hampered their proper use: (1) Word embedding mixes multiple senses and biased uses of a word in the same representation through superposition, and (2) Word embedding, though successful in capturing equivalences and analogies, cannot be used for an arbitrary purpose and combined using typical vector operations because the latter does not have a proper semantic basis. Domain-specific (or semantic) embeddings mitigate the ambiguity issues somewhat but are at the mercy of the domain-specific corpus being used.

In the past, we decried syntactic approaches as being limited and proposed semantics-enhanced approaches as providing a reliable basis for intelligent behavior. Now, "syntactic approaches" have resurfaced in more fancy forms where word embeddings, which have adequately captured the contextual meaning of words in isolation, have been combined in ways not warranted by their semantics. What does adding and averaging two-word embeddings truly signify? How do we know that the "mixture" word embedding associated with a sentence, obtained by an indiscriminate combination of individual word embeddings associated with words in a sentence, captures the intended semantics of the sentence? In fact, in all likelihood, it may end up obscuring important distinctions because it is unclear if the semantics is compositional when using word embeddings. For example, word embeddings satisfy relationships such as Man + Queen = Woman + King, and Man :: Woman :::: King :: Queen. However, in theory, it is also possible to have word embeddings A + B = C + D = E, where sentence A … B and C … D may not be semantically related. Similarly, one cannot combine 300-dimensional word embeddings obtained for a word through the analysis of different text corpora that provide orthogonal perspectives on the meaning (such as generic vs domain-specific). And because these word embeddings are inscrutable and their combinations uninterpretable, how can we assess the robustness of these techniques for new datasets, or argue about the inadequacy/completeness of a validation suite? How can we make "linguistic intelligence" intelligible? At this juncture, the best we can do is to learn a representation for all the words from the same corpus so that at least they can be compared on compatible dimensions.

Alternatively, we are permitted to concatenate multiple word embeddings associated with a word that captures different perspectives, as opposed to adding them, so that the perspectives can be kept independent. Dot-product operations can still make sense for measuring similarity in both these settings because each dimension of the (extended) vector is still compatible. However, we may need to determine separately the emphasis to be placed on each perspective to get an empirically satisfactory global model. Further, in this setting, it is unclear how to separate the multiple senses. Thus, the problem of finding word embedding vector operations that can be justified on semantic-grounds is still open, albeit with some notable but limited success for word equivalences. In these situations, finding a means to incorporate declarative linguistic knowledge for disambiguation and improving coverage can improve text analytics.

**HYBRIDIZING NEURAL NETWORKS AND DEEP LEARNING WITH KNOWLEDGE-BASED REASONING SYSTEMS**

The deep learning models such as GPT3, VGGNet, GNMT, BERT, and XLNet use billions of parameters and may overfit training data in ways that may run the risk of a critical failure that may surface only by chance and at an inopportune time. For instance, some of the GPT3 generated examples seem like a locally sensible but globally incoherent collage of sentence fragments. And if we train these systems on text available on the Web without careful oversight, in due course, machine-generated text may end up contaminating these deep learning systems in ways that will be counter-productive such as by further reinforcing existing biases in the corpus. **While we share Hinton's optimism in the untapped potential of deep learning (http://j.mp/3c7xYPg), we also strongly believe that one way to achieve novel breakthroughs is by integrating knowledge with deep learning in a variety of ways by infusing knowledge into deep learning** [Sheth et al 2019].

**Intelligent Data Processing and Types of Knowledge**

In the tasks that primarily are about processing and analyzing a large amount of data to classify, predict and otherwise learn or derive insight from, deep learning systems have increasingly beaten human-level performance. Now the AI systems are moving towards continuous interactions with the humans (i.e., human-centered and human-in-the-loop AI), and in the process utilize multisensory and multimodal data. They also call for an AI system to be mindful of humans they interact with (e.g., show empathy) and exhibit human-like intelligence and flexibility. Such applications require contextualization (w.r.t. the external environment), personalization (w.r.t. the user) and abstractions (w.r.t. different levels of detail and purpose) [Sheth et al 2018]. **The Holy Grail is to develop interpretable systems whose internal operations are transparent enough to feel confident about a system's expected behavior and provide an explanation interface that communicates the outcomes and the underlying rationale in ways that inspire confidence in the end-user of the system for wider and sustained adoption. Data and statistical AI techniques are increasingly inadequate for these capabilities, and knowledge becomes increasingly important.**

**Knowledge comes in many different forms, and different forms of knowledge can play a fundamental role in mediating activities involving different levels of abstractions. The knowledge may be at different levels of representational expressiveness or richness.** Figure 2 demonstrates stratified knowledge capturing low-level representation relevant to the data processing to the rich domain and application-level semantics and abstractions (for example, for natural language understanding). In a particular pipeline, these types of knowledge may be used (or infused using statistical or deep learning techniques) in a different order than the one shown here, depending on different types of data (e.g., clinical text, social media data, sensor/IoT), application domains (e.g., for mental health we may use knowledge derived from DSM-5 while for

addiction, we may use Drug Abuse Ontology), and application purpose (in medicine, for diagnosis, treatment, prediction, patient self-management, etc.).

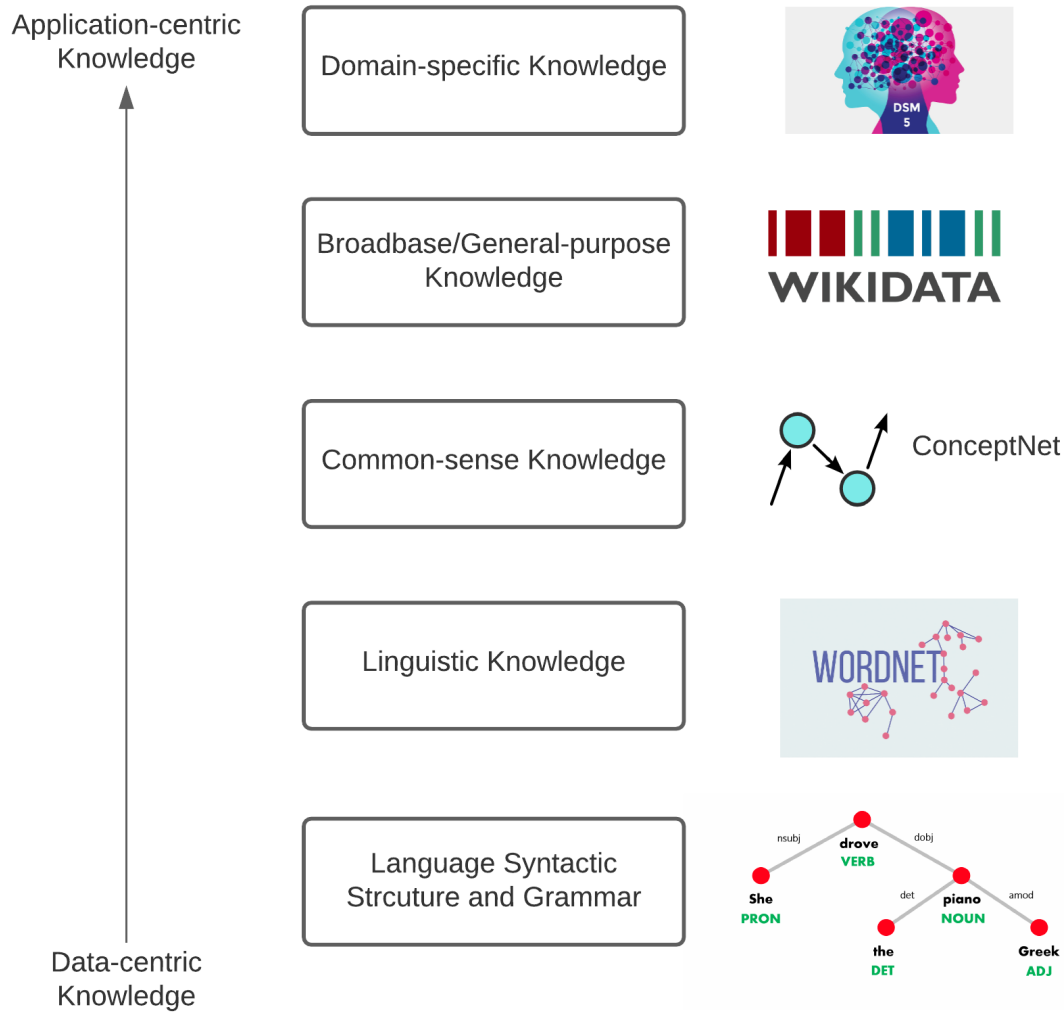

Figure 2: Stratified knowledge types with one example knowledge that can be directly used or extracted from, relevant to natural language understanding. At any level, there can be more than one knowledgebase (for example, in addition to Wikidata, for an application that requires geographic knowledge, we may use OpenStreetMap as a source of relevant knowledge).

**Abstraction and Analogies**

One of the key issues underlying the debate about the effectiveness of data-driven approaches vs knowledge-based approaches is the source and the creator of abstractions [Mitchell 2021]. Data-driven approaches derive abstractions utilizing the trends and patterns explicit or implicit in the copious amounts of data, and its utility is tied to the representativeness of the data for the domain of discourse. In contrast, knowledge-based approaches typically present abstractions that are obtained directly or indirectly by end-user applications and goals. Insofar as these two orthogonal approaches yield overlapping abstractions, we can use them synergistically. Further, their hybridization presents a viable strategy to leverage their complementary strengths. Data-driven abstractions can group values into coarser categories based on data usage patterns but labeling them sensibly and developing abstractions with specific purpose usually requires human-in-the-loop. For instance, abstractions of temperature ranges and temperature thresholds that are relevant to different states of matter of $H_2O$ (such as ice, melting point, water, boiling point, and steam) are different from those relevant to human health (such as normal temperature, fever, mild fever, high fever, and chill). In circuit design, we can describe a full adder at different levels of abstraction in terms of two half-adders, or using logic gates (either using AND, OR, and NOT gates, or NAND/NOR gates), or using flip-flop circuits, or using transistor netlists, etc. In the context of image recognition and transfer learning, the later stages of neural network layers capture higher-level abstractions/features that mirror the nature/components of the images in the training data and the end-user application. Further, probing these features can not only improve our confidence in its working but also provide insights about the potential limitations that can be addressed and remedied. Our effort in deep knowledge-infused learning (within shallow, semi-deep, and deep infusion variety [Sheth et al 2019]) is based on this intuition. Analogical reasoning involving primitive shapes (e.g., circle, square, and triangle) and their relationships (e.g., inside, outside, adjacent, above, below, left, right, and overlapping) in an image requires non-trivial human insights and geometric reasoning [Evans 1987]. This example presents an interesting situation where we can use deep learning algorithms to recognize primitive shapes and then use geometric analogical reasoning to solve non-trivial puzzles (e.g., geometric analogy problems that appear on intelligence tests).

As discussed above, there are several different motivations and applications for combining deep-learning powered neural networks with knowledge-based systems. **Developing a modular framework to combine data-driven approaches that are effective at extracting useful features from low-level sensory and linguistic data, with declarative models of normal behavior such as through the incorporation of physical laws, medical knowledge, linguistic and domain-specific knowledge, as well as anticipated failure modes and exceptions, can go a long way in leveraging insights from and exploiting the duality of data and knowledge.**

**CONCLUSION**

Knowledge representation as expert system rules or using frames and a variety of logics played a key role in capturing explicit knowledge during the hay days of AI in the past century (also referred to as the first wave of AI). Such knowledge aligned with planning and reasoning is part of what we refer to as Symbolic AI. The resurgent AI of this century in the form of Statistical AI (also referred to as the second wave of AI) has benefitted from massive data and computing. On some tasks, deep learning methods have even exceeded human performance levels. This gave the false sense that data alone is enough, and explicit knowledge is not needed ([Kambhampati 2021], also http://bit.ly/ExpKnow). But as we start chasing machine intelligence that is comparable with human intelligence, there is an increasing realization that we cannot do without explicit knowledge.

**Neuroscience (role of long term memory, strong interactions between different specialized regions of data on tasks such as multimodal sensing), cognitive science (bottom brain versus top brain, perception versus cognition), brain-inspired computing, behavioral economics (Kahneman's thinking fast vs thinking slow, or system 1 versus system 2), and other disciplines point to need for furthering AI to neuro-symbolic AI (i.e., hybrid of Statistical AI and Symbolic AI, also referred to as the third wave of AI). As we make this progress, the role of explicit knowledge becomes more evident, especially in our endeavor to support human-like intelligence, our desire for AI systems to interact with humans naturally, and our need to explain the path and reasons for AI systems' workings. Further, the nature of knowledge needed to support understanding and intelligence is varied and complex, ranging from linguistic, common sense, and general (world model) to specialized (e.g., geographic) and domain-specific (e.g., mental health)**. Despite this complexity, such knowledge can be created and maintained at scale (even dynamically and continually).

We are exploring knowledge-infused learning as an example strategy for fusing statistical and symbolic AI in a variety of ways to realize our vision. We believe that symbolic explicit knowledge is an indispensable substrate to integrate and abstract multi-modal data to build an intelligent system that can meaningfully interact with and assist humans and help navigate through the intricacies and nuances of the real-world scenarios (that have been realized, predicted, or even imagined) to make decisions and obtain desired outcomes.

About the Authors:

Amit Sheth (http://amit.aiisc.ai) is the founding director of the Artificial Intelligence Institute at the University of South Carolina (http://aiisc.ai). He is a fellow of IEEE, AAAI, AAAS, and ACM. He received his BE in EEE (Hons) from B.I.T.S. Pilani, India in 1981, and MS and Ph.D. in Computer & Information Sciences from the Ohio State University in 1983 and 1985, respectively. His current foundational AI interests are in knowledge-infused learning and explicitly, knowledge-enabled neuro-symbolic computing, and translational AI interests involve applications to many domains including personalized digital health, public health, epidemiology, biomedicine, smart manufacturing, social good, and education.

Krishnaprasad Thirunarayan (http://cecs.wright.edu/~tkprasad/). received a B. Tech. degree in electrical engineering from the Indian Institute of Technology, Madras in 1982, an M. E. degree in computer science from the Indian Institute of Science, Bangalore in 1984, and a Ph.D. degree in computer science from the State University of New York at Stony Brook, Stony Brook in 1989. He is currently a Professor in the Department of Computer Science and Engineering at Wright State University, Dayton, Ohio. His research interests are in semantic analysis of social, sensor and text data, knowledge representation and reasoning, artificial intelligence and machine learning, and trust networks.